\documentclass[letterpaper,10pt,conference]{ieeeconf}

\IEEEoverridecommandlockouts
\usepackage{cite}
\usepackage{amsmath,amssymb}
\usepackage{booktabs,multirow}
\usepackage{graphicx}
\usepackage{xcolor}
\usepackage{url}
\usepackage{etoolbox}
\usepackage{placeins}
\usepackage[hidelinks]{hyperref}
\hypersetup{
  pdftitle={Execution-Aligned Progressive Noise for Consistent Asynchronous Replanning in Generative Robot Policies},
  pdfauthor={Di Wu, Ping Liu, Xuhua Chen, He Zheng, Lingfeng Zhang, Tao Zhang}
}
\makeatletter
\patchcmd{\thebibliography}{\section*{References}}{\section*{REFERENCES}}{}{}
\makeatother
\allowdisplaybreaks

\newcommand{\R}{\mathbb{R}}

\newcommand{\N}{\mathcal{N}}
\newcommand{\Cov}{\operatorname{Cov}}

\newcommand{\norm}[1]{\left\lVert #1\right\rVert}
\newcommand{\piHalf}{\ensuremath{\pi_{0.5}}}

\title{\LARGE \bf
Execution-Aligned Progressive Noise for Consistent Asynchronous Replanning in Generative Robot Policies
}

\author{Di Wu$^{1,2,*}$, Ping Liu$^{1,*}$, Xuhua Chen$^{1}$, He Zheng$^{1}$, Lingfeng Zhang$^{1}$, and Tao Zhang$^{1,\dagger}$%
\thanks{$^{1}$Magiclab Robotics Technology Co., Ltd., China.}%
\thanks{$^{2}$Southeast University, Nanjing, China.}%
\thanks{$^{*}$These authors contributed equally.}%
\thanks{$^{\dagger}$Corresponding author.}%
}

\begin{document}

\maketitle
\thispagestyle{empty}
\pagestyle{empty}

\begin{abstract}
Continuous asynchronous replanning is essential for real-time generative robot policies, but independent stochastic initialization can cause mode switching and inconsistent continuation across action chunks. We propose Execution-Aligned Progressive Noise (EAPN), which introduces structured stochasticity at both inter-chunk and intra-chunk levels. Across replanning steps, EAPN propagates a shared noise trajectory and aligns it with the actual execution displacement, establishing execution-aligned inter-chunk correlation. Within each action chunk, it models temporal correlation along action time. The aligned stochastic history is further combined with committed
action context to condition subsequent generation, allowing new chunks to continue from execution-consistent generative states rather than restart from independent noise. We evaluate EAPN on D3IL, Kinetix, LIBERO, and real-world manipulation tasks. EAPN improves multimodal behavior consistency on D3IL and achieves an average success rate of 88.59\% on Kinetix. On LIBERO, it remains robust and maintains strong task performance even under long inference delays. Real-robot experiments further achieve 90.0\% success on Object Storage and 96.7\% on bimanual Cloth Folding, demonstrating reliable continuous execution under asynchronous replanning.
\mbox{\textbf{Project Page:}
\href{https://embodied.magiclab.top/works/eapn/index.html}{\textcolor{blue}{\textit{EAPN}}}}
\end{abstract}

\section{INTRODUCTION}
\label{sec:introduction}

Generative models, including Flow Matching, support robot decision-making and action-sequence generation~\cite{lipman2023flowmatching,chi2023diffusionpolicy,black2024pi0,physicalintelligence2025pi05}.
Unlike deterministic regression policies, they can model multimodal conditional distributions over plausible future motions.
Their iterative generation, however, operates
at a lower frequency than closed-loop control.
To bridge this gap, generative robot policies commonly use Action Chunking: the policy predicts a future action sequence at each call~\cite{zhao2023act,chi2023diffusionpolicy}. High-latency Vision-Language-Action (VLA) models and World-Action Models further use asynchronous execution, generating the next chunk while the current one is still being executed~\cite{black2025rtc,motubrain2026wamrealtime}. Because the robot continues moving during inference, the physical execution state has already advanced when the new prediction becomes available, causing prediction-execution misalignment, stale actions, and discontinuities between adjacent chunks.

Existing methods mainly address this issue through action-space constraints and execution-time alignment. RTC and
Training-Time RTC (ttRTC) condition generation on a committed action prefix~\cite{black2025rtc,black2025trainingrtc}; VLASH and FutureRTC estimate the execution-time context at which a new prediction will actually take effect~\cite{tang2025vlash,jiang2026futurertc}; and Legato, SEAM, and Soft RTC improve continuation through learned dynamics, velocity guidance, or soft action priors~\cite{liu2026legato,zhan2026seam,liu2026softrtc}. These approaches improve temporal consistency, but their constraints concentrate on the executed or imminently executed trajectory prefix. Beyond this explicitly constrained region, the remaining suffix is still largely determined by a newly sampled stochastic source. Stochastic-state dependency across consecutive generation processes therefore remains insufficiently modeled.

Most generative policies initialize every inference independently from Gaussian noise. For a multimodal conditional distribution, different stochastic initializations can drive temporally overlapping inferences toward different, yet individually high-probability, local modes. Thus, even when the committed prefix is continuous, the freely generated suffix may gradually deviate from the behavioral mode represented by the previous inference and switch to a trajectory with a different motion intent or manipulation strategy. Such cross-inference mode switching need not appear as a large single-step error and is therefore difficult to identify from local trajectory smoothness alone. In long-horizon manipulation, however, inconsistencies in motion direction, end-effector state, contact relation, or manipulation semantics can accumulate across successive Action Chunks and cause substantial execution deviations. This motivates modeling not only temporal alignment in action space, but also stochastic-state continuity across consecutive replanning steps.

\begin{figure*}[t]
    \centering
    \includegraphics[width=0.98\textwidth]{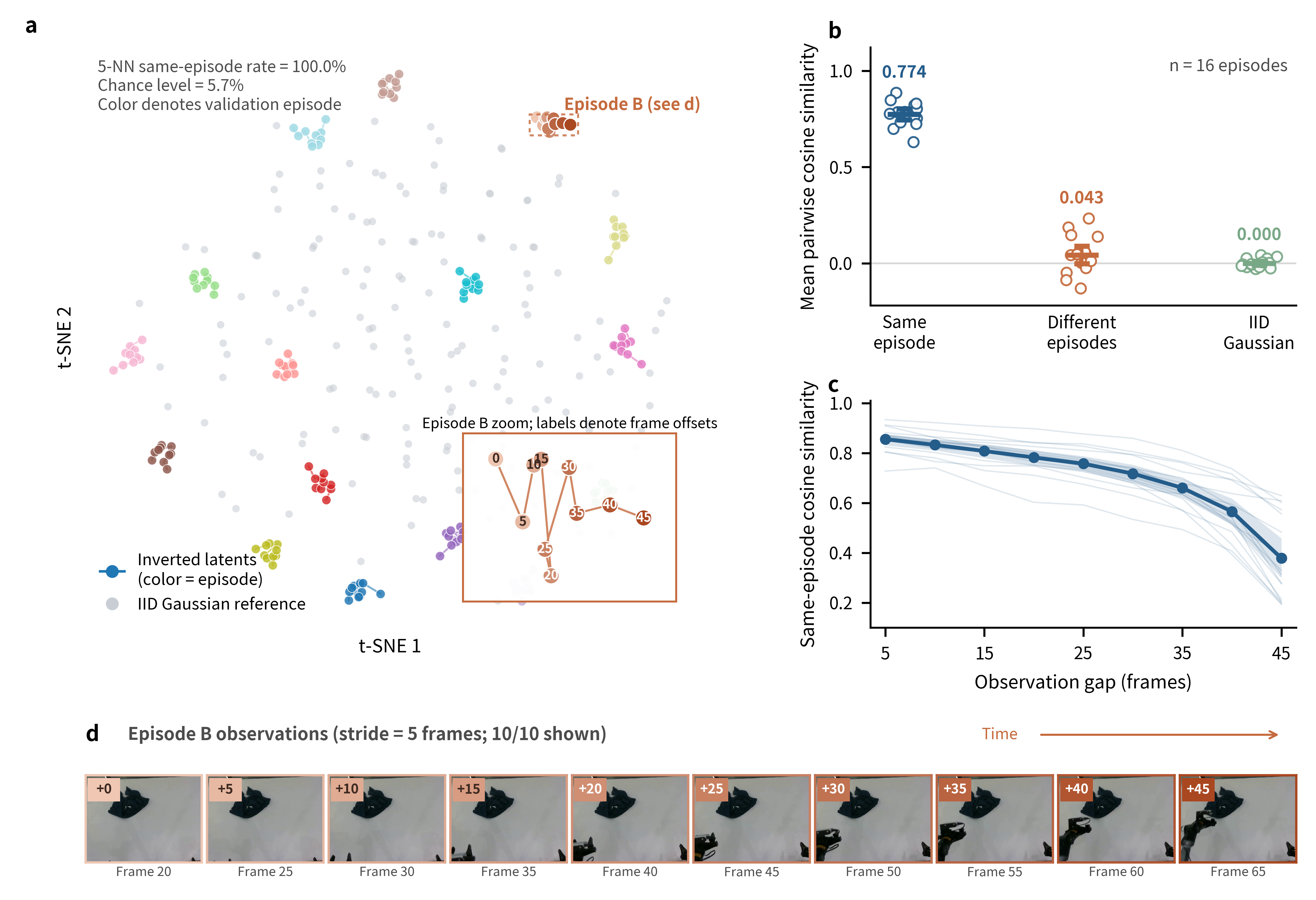}
    \caption{Frame-interval inversion of a standard \piHalf action policy reveals temporally local dependency in the initial noise. The policy is trained with token-wise independent Gaussian noise, and the inversion procedure contains no EAPN structure. \textbf{(a)} Execution-aligned inverted initial noise from 16 held-out validation episodes together with an independent Gaussian reference; the same-episode 5-NN hit rate in the raw space is 100\%. \textbf{(b)} Mean cosine similarity is 0.774 within the same episode, 0.043 across different episodes, and 0.000 for the independent Gaussian reference. \textbf{(c)} As the observation interval increases from 5 to 45 frames, the within-episode similarity decreases continuously from 0.855 to 0.378. Thick curves denote the mean over 16 episodes and shaded regions indicate 95\% bootstrap confidence intervals. \textbf{(d)} Ten real observations from frames 20--65 of Episode B are matched one-to-one with their inverted latent points.}
    \label{fig:stride5_inversion}
\end{figure*}

Figure~\ref{fig:stride5_inversion} provides empirical evidence for this view. Flow inversion of consecutive action windows reveals pronounced temporal locality in source latents within the same episode: correlations gradually decay as the temporal separation increases, whereas correlations across different episodes remain near zero. Although the base policy is trained with independent Gaussian sources, temporally continuous behaviors therefore exhibit stable cross-time latent dependency. This suggests that preserving execution-consistent stochastic relations across neighboring inferences may provide a useful prior for continuous replanning.

Motivated by this observation, we propose
\emph{Execution-Aligned Progressive Noise} (EAPN),
a stochastic-state dependency modeling method for
continuous generative replanning. Rather than directly
constraining deterministic action outputs, EAPN introduces structured stochasticity at two complementary levels.
First, \emph{execution-aligned inter-chunk correlation} propagates stochastic states across consecutive policy
calls according to the actual execution displacement, so that temporally corresponding positions in neighboring Action Chunks share consistent stochastic context.
Second, \emph{intra-chunk temporal correlation} models dependency along action
time within each chunk. Together, these components provide a temporally structured stochastic prior spanning both inference events and action time, reducing stochastic drift and unnecessary mode switching while remaining compatible with asynchronous execution. Our contributions are:

\begin{itemize}
    \item \textbf{Execution-aligned latent correlation.} Through Flow inversion, we show that neighboring Action Chunks exhibit correlated initial noise whose peak shifts with the actual execution displacement and decays as the window separation increases.

    \item \textbf{Joint inter- and intra-chunk stochastic modeling.} EAPN recursively propagates noise
    across consecutive inferences while introducing temporal correlation within each Action Chunk, reducing stochastic drift and unnecessary mode switching caused by independent reinitialization.

    \item \textbf{Evaluation across simulation and real robots.}
    Experiments on D3IL, Kinetix, LIBERO, and two real-world tasks demonstrate improved task performance, multimodal consistency, cross-chunk continuity, and robustness to inference delays, achieving 90.0\% on Object Storage and 96.7\% on bimanual Cloth Folding.
\end{itemize}

\section{RELATED WORK}
\label{sec:related}

\subsection{Asynchronous Action-Chunk Execution and Temporal Alignment}
Action Chunking reduces policy invocation frequency but introduces prediction--execution misalignment when inference overlaps continuous control~\cite{zhao2023act,chi2023diffusionpolicy}. RTC conditions generation on a committed action prefix, while Training-Time RTC learns this continuation during training~\cite{black2025rtc,black2025trainingrtc}. REMAC handles discrepancies between planned and executed actions through masked chunks~\cite{wang2026remac}; VLASH and FutureRTC improve alignment using future-state rollout or execution-time state prediction~\cite{tang2025vlash,jiang2026futurertc}; DiscreteRTC exploits discrete-diffusion inpainting~\cite{wang2026discretertc}; and Action ControlNet introduces a lightweight delay-aware residual adapter~\cite{guo2026actioncontrolnet}. Related delay and chunk-overlap issues also arise in asynchronous World--Action Models~\cite{motubrain2026wamrealtime}. These methods primarily align historical states/actions with newly generated trajectories.

\subsection{Cross-Chunk Continuation and Multimodal Consistency}
A complementary line of work targets continuation quality across chunks. BID selects candidates using backward coherence and forward contrast~\cite{liu2025bid}, while Self-Guided Action Diffusion reuses the previous decision as sampling guidance~\cite{malhotra2025selfgad}. Legato learns continuation dynamics for multimodal Flow Matching~\cite{liu2026legato}; SEAM uses velocity guidance derived from the previous chunk~\cite{zhan2026seam}; and Soft RTC replaces a hard prefix with an editable action prior~\cite{liu2026softrtc}. ChunkFlow, FocalPolicy, and POTR further improve chunk-boundary or long-horizon coherence through seam-aware objectives, frequency-aware modeling, or constrained sampling-time guidance~\cite{yang2026chunkflow,he2026focalpolicy,fang2026potr}. These approaches encourage consistency through candidate selection, action/velocity guidance, overlap constraints, or learned continuation dynamics. EAPN instead asks whether the stochastic state from which each generation begins should itself remain temporally related across replanning events, providing a complementary mechanism to action-space continuation.

\subsection{Source Priors and Correlated Stochastic Initialization}
Beyond robot policies, PYoCo uses a correlated noise prior in video diffusion models to preserve temporal structure across frames~\cite{ge2023pyoco}, showing that source correlation itself can support temporal consistency. Recent work has also revisited source distributions for generative robot policies. Streaming Flow Policy initializes near the previous action~\cite{jiang2025streaming}; Action-to-Action Flow Matching, WarmPrior, and Temporal Policy construct history-informed sources from recent actions or proprioception~\cite{jia2026a2a,kang2026warmprior,miller2026temporalpolicy}; and LeaP learns source mean and variance from the current proprioceptive state~\cite{dai2026leap}. PAINT uses Flow inversion to select initial noise compatible with a historical prefix~\cite{ho2026paint}. These studies establish the generative starting point as an important design dimension. Unlike PYoCo's fixed correlation structure for video generation, EAPN models a persistent, execution-aligned dependency across consecutive robot replanning windows while simultaneously retaining structured temporal correlation within each chunk.

\section{METHOD}
\label{sec:method}

\subsection{Problem Formulation and Method Overview}
\label{subsec:problem}
At inference step $k$, a flow-based policy generates an $H$-step action sequence from observation $o_k$:
\begin{equation}
    A_k=[a_{k,1},\ldots,a_{k,H}]\in\R^{H\times D_a}.
\end{equation}
Standard Flow Matching~\cite{lipman2023flowmatching} starts from source noise $Z_k\in\R^{H\times D_a}$ and follows
\begin{equation}
    X_{k,t}=(1-t)Z_k+tA_k,
    \qquad U_k=A_k-Z_k
\label{eq:fm_path}
\end{equation}
EAPN adopts the committed-prefix formulation of Training-Time RTC~\cite{black2025trainingrtc}: actions already committed for execution remain fixed while only the future suffix is regenerated. Beyond action conditioning, EAPN explicitly preserves stochastic-state continuity across consecutive inferences through an execution-aligned structured source and historical-noise conditioning.

\subsection{Execution-Aligned Structured Noise}
\label{subsec:structured_noise}
Because consecutive asynchronous prediction windows shift along physical time,
stochastic states should be aligned by actual execution progress rather than by identical token indices.
EAPN therefore models both inter-chunk execution-aligned dependency and intra-chunk temporal dependency.
The structured source can be written using a shared AR(1) process $G_t$, persistent across policy calls,
and a private within-window AR(1) process $U_{k,h}$, Here, AR(1) denotes a first-order
autoregressive process,
where each state depends on the previous state with Gaussian noise:
\begin{equation}
    G_t=\rho G_{t-1}+\sqrt{1-\rho^2}\,\epsilon_t^G,
    \qquad \epsilon_t^G\sim\N(0,I),
\label{eq:G_ar1}
\end{equation}
\begin{equation}
    U_{k,h}=\rho U_{k,h-1}+\sqrt{1-\rho^2}\,\epsilon_{k,h}^U,
    \qquad \epsilon_{k,h}^U\sim\N(0,I),
\label{eq:U_ar1}
\end{equation}

The private process is recursively generated along the action-time dimension, introducing temporal correlation between noise tokens within each Action Chunk.
with source token
\begin{equation}
    z_{k,h}
    =\sqrt{\gamma}\,G_{T_k+h}
    +\sqrt{1-\gamma}\,U_{k,h}.
\label{eq:structured_noise}
\end{equation}
Here, $\rho\in[0,1)$ controls temporal correlation decay and $\gamma\in[0,1]$ balances shared history against window-specific innovation.

\begin{figure*}[t]
\centering
\includegraphics[width=0.98\textwidth]{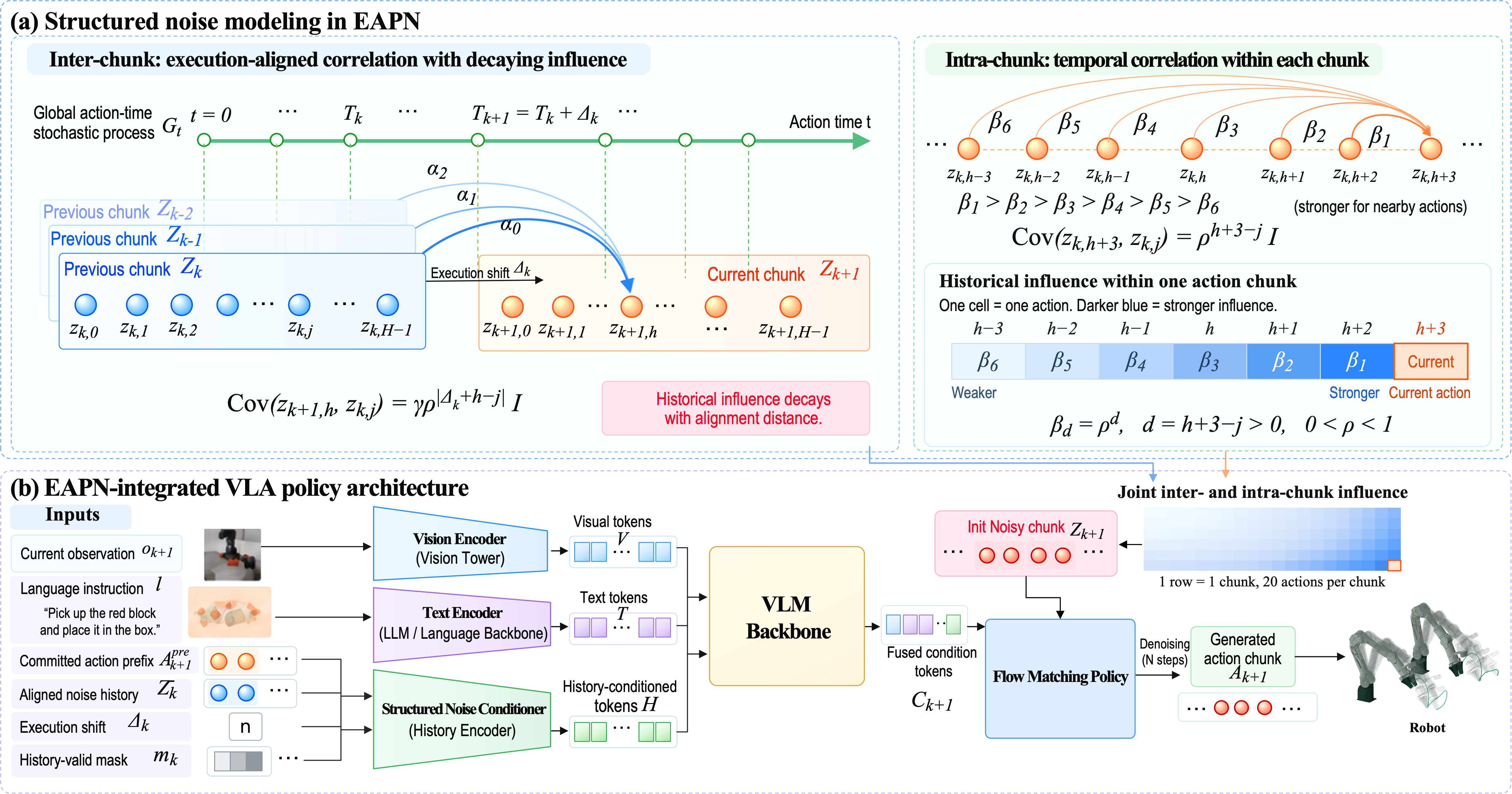}
\caption{Overview of the EAPN method.}
\label{fig:eapn_overview}
\end{figure*}

Both AR(1) processes have standard Gaussian stationary marginals, hence
\begin{equation}
    z_{k,h}\sim\N(0,I).
\label{eq:noise_marginal}
\end{equation}
Within a chunk, correlation decays with temporal distance. Across chunks, re-indexing the shared trajectory by the actual window displacement $\Delta_k$ gives
\begin{equation}
    \Cov(z_{k+1,h},z_{k,j})
    =\gamma\rho^{|\Delta_k+h-j|}I.
\label{eq:inter_cov}
\end{equation}

Equation~\eqref{eq:inter_cov} characterizes how the stochastic dependency between consecutive Action Chunks is aligned with the actual execution displacement.
Thus, the strongest dependency occurs near the execution-aligned position $j\approx h+\Delta_k$, rather than
at a fixed same-index location. Equation~\eqref{eq:noise_marginal} preserves only the single-token Gaussian marginal; the joint covariance of the full $H\times D_a$ source changes. The complete EAPN model is therefore trained to adapt to this structured joint source rather than treated as a training-free distribution-preserving sampler.

\subsection{History-Conditioned Noise Representation}
\label{subsec:conditioner}
Correlated noise alone does not tell the policy which historical stochastic state corresponds to the current window. Given the previous source $Z_{k-1}$, EAPN aligns it using the actual execution displacement:
\begin{equation}
    \bar Z_{k-1}
    =\mathcal A(Z_{k-1},\Delta_{k-1}),
\label{eq:noise_align}
\end{equation}
where $\mathcal A(\cdot)$ shifts only tokens that retain temporal correspondence after execution and a validity mask $m_k$ marks positions for which no historical counterpart exists. This distinction is important at episode starts and whenever the previous stochastic history is unavailable. The conditioner computes
\begin{equation}
    c_k^{\mathrm{hist}}
    =\mathcal C_\phi
    \left(
        \bar Z_{k-1},
        \Delta_{k-1},
        D_k,
        m_k
    \right),
\label{eq:conditioner}
\end{equation}
and injects this representation into the policy. Here, $\Delta_{k-1}$ records how many steps of the previous stochastic trajectory have been consumed, $D_k$ is the current inference delay/committed-prefix length, and $m_k$ distinguishes valid history from first-inference or history-invalid states. EAPN therefore learns
\begin{equation}
    v_\theta
    \left(o_k,X_{k,t},\tau_{k,t},c_k^{\mathrm{hist}}\right),
\label{eq:history_conditioned_velocity}
\end{equation}
where token-wise flow times $\tau_{k,t}$ distinguish the fixed prefix from the generated suffix; committed actions are written directly into $X_{k,t}$. This makes the aligned stochastic history an explicit policy condition rather than an implicit sampler modification.

\subsection{Joint Training and Asynchronous Deployment}
\label{subsec:joint_inference}
During training, delay $D$ and displacement $\Delta$ are sampled from the evaluation support, and adjacent source pairs $(Z^{-},Z)$ with matching execution geometry are generated through the shared process $G$. The aligned $Z^{-}$ conditions the current sample. To handle episode starts or invalid history, the entire history is dropped with probability $p_{\mathrm h}$; the current source is then resampled independently and both the aligned history and its validity mask are zeroed. This prevents the model from assuming that historical stochastic context is always available. For committed positions $h\leq D$, we set $(X_h,\tau_h)=(A_h,1)$, while the suffix follows Eq.~\eqref{eq:fm_path}. The Flow Matching loss is applied only to the generated suffix:
\begin{equation}
    \mathcal L_{\mathrm{FM}}
    =\frac{1}{H-D}\sum_{h>D}
    \norm{
        v_{\theta,h}(o,X,\tau,c^{\mathrm{hist}})
        -(A_h-Z_h)
    }_2^2.
\label{eq:masked_fm_loss}
\end{equation}

At deployment, an episode reset clears the action buffer and disables history. The first inference samples a new shared/private AR(1) window. Thereafter, the shared trajectory advances by the number of actions actually consumed, the previous source is aligned by the same $\Delta_k$, and the resulting $c_k^{\mathrm{hist}}$ conditions every flow step while the first $D_k$ actions remain fixed. After generation, the current shared trajectory and source are stored for the next replanning step. EAPN leaves the base Flow Matching velocity-field objective and the original action-chunk execution logic unchanged; it modifies only the joint temporal source structure and adds a conditioning representation of aligned stochastic history. Importantly, $D_k$ controls the committed prefix, whereas $\Delta_k$ controls stochastic-state advancement and alignment; under a fixed execution period they may coincide, but this equality is not required by EAPN.

\section{EXPERIMENTS AND RESULTS}
\label{sec:experiments}

\subsection{Inversion Analysis of a Standard Policy}
We first test whether consecutive Action Chunks
exhibit latent correlation even when the base
policy is trained with independent Gaussian sources.
Using \piHalf~\cite{physicalintelligence2025pi05}, whose training independently samples Gaussian sources at every action-time position, we select 16 held-out episodes, collect overlapping chunk pairs with actual execution displacements $D=1,\ldots,9$, and invert their action sequences into noise space with the trained conditional vector field. Because no cross-inference coupling is imposed during training, any recovered temporal structure provides direct evidence about stochastic dependency associated with consecutive behavior.

\begin{figure*}[!t]
    \centering
    \includegraphics[width=0.99\textwidth]
    {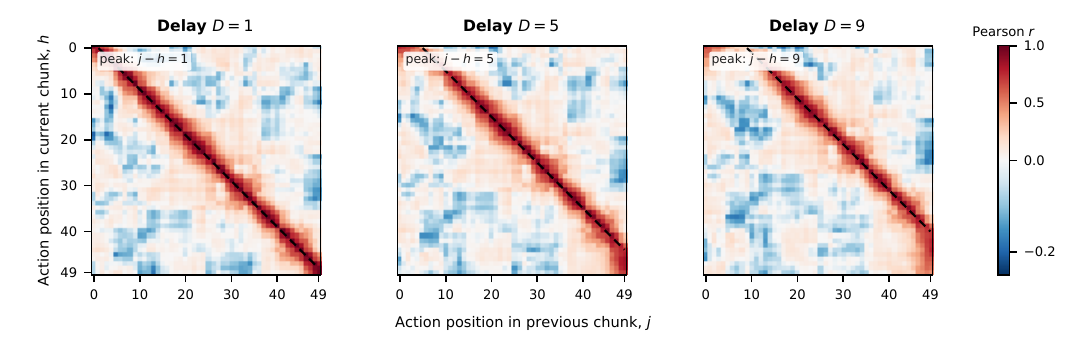}
    \caption{Execution-aligned latent structure in a standard \piHalf policy.
    Latent correlation matrices between neighboring Action Chunks for actual execution displacements $D=1,5,9$.
    The black dashed line indicates the execution-aligned position $j=h+D$ determined by the actual execution displacement.}
    \label{fig:inverted_alignment}
\end{figure*}

Figure~\ref{fig:inverted_alignment} shows that, for $D=1,5,9$, the correlation peak follows the execution-aligned relation $j-h=D$ rather than a fixed same-index correspondence. The location of the strongest correlation therefore moves with the actual execution displacement, rather than remaining tied to a particular token index. This pattern is consistent across the representative delays and indicates that physically overlapping chunks preserve a stable execution-aligned structure in inverted noise space. The observation provides direct empirical support for using the actual displacement to define cross-chunk covariance in Eq.~\eqref{eq:inter_cov}.

\subsection{D3IL: Multimodality Preservation and Cross-Inference Mode Consistency}
D3IL provides interpretable multimodal behavior descriptors~\cite{jia2024d3il}.
We use its Avoiding task as a mechanism diagnostic because different obstacle-avoidance paths
correspond to clear behavioral modes. For matched initial states, we perform paired rollouts with
IID noise and EAPN using DDPM-ACT~\cite{chi2023diffusionpolicy,zhao2023act}, which combines stochastic
diffusion generation with action chunking. Besides task success, we measure the Mode Switch Rate (MSR),
the fraction of adjacent classifiable replanning pairs that change mode within the same obstacle stage;
normal progression between stages is excluded. Switch Gap is the $\ell_2$ distance
between adjacent chunks at the actual switching position and captures local replanning discontinuity.

\begin{table}[t]
\centering
\caption{Task success and behavioral consistency under consecutive replanning on D3IL Avoiding.}
\label{tab:d3il_mode}
\setlength{\tabcolsep}{3.0pt}
\begin{tabular}{@{}lccc@{}}
\toprule
Method & Success $\uparrow$ & MSR $\downarrow$ & Switch Gap $\downarrow$ \\
\midrule
DDPM-ACT & 73.13\% (351/480) & 3.969\% & 0.001913 \\
\textbf{EAPN} & \textbf{78.96\% (379/480)} & \textbf{3.167\%} & \textbf{0.001672} \\
\bottomrule
\end{tabular}
\end{table}

EAPN increases success from 73.13\% to 78.96\%, a gain of 5.83 percentage points, while reducing MSR from 3.969\% to 3.167\% and Switch Gap from 0.001913 to 0.001672. Because the alternative paths in Avoiding have direct geometric interpretations, these changes indicate fewer cross-inference jumps between valid behavior modes and smoother executed transitions. Importantly, the improvement is obtained without collapsing the policy to a single behavior, suggesting that EAPN stabilizes mode continuation while preserving multimodal capacity.

\subsection{Kinetix: Robustness to Dynamic Inference Delay}
\begin{figure*}[!t]
    \centering
    \includegraphics[width=0.99\textwidth]{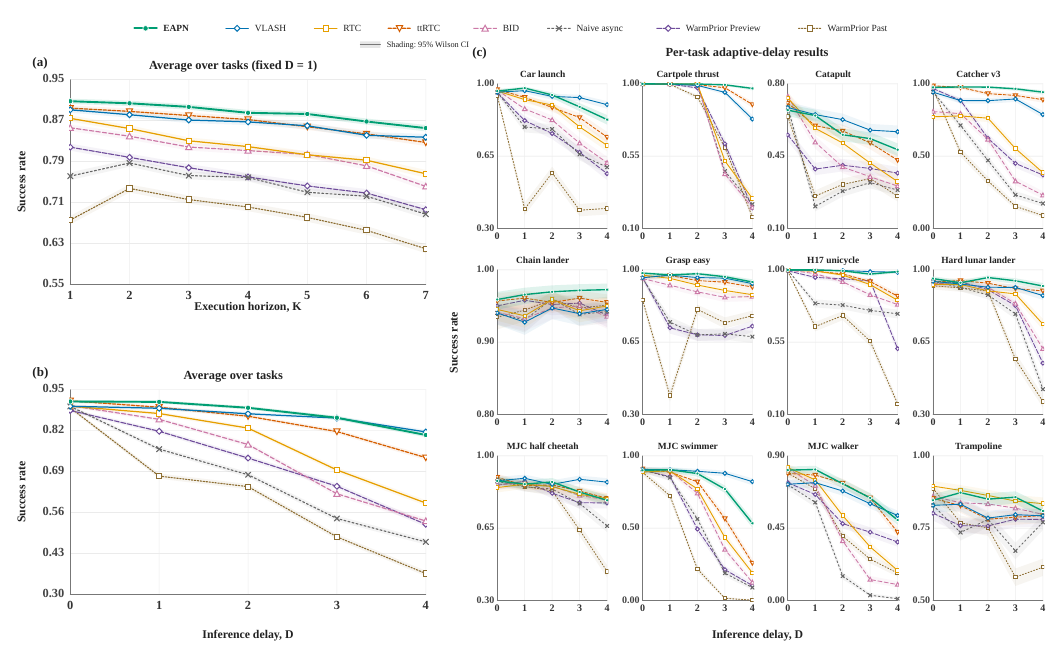}
    \caption{Comparison under asynchronous inference on Kinetix. Left: average success rate over 12 tasks for different execution horizons $K$ with fixed $D=1$, and for different inference delays $D$. Right: per-task results.}
    \label{fig:kinetix_main_results}
\end{figure*}

Kinetix is an open-ended 2D physics-control benchmark built on Jax2D~\cite{matthews2025kinetix}. Following the asynchronous protocol introduced with RTC on this benchmark~\cite{black2025rtc}, we set the prediction horizon to $H=8$ and vary the committed/inference-prefix length as
\begin{equation}
    D\in\{0,1,2,3,4\}.
\end{equation}
For each delay, we evaluate multiple execution horizons $K$ over which actions must be supplied continuously and report both the 12-task average and per-task results. Figure~\ref{fig:kinetix_main_results} shows the expected degradation as either $K$ or $D$ increases, but EAPN declines more gradually and remains stronger on most tasks. This indicates that execution-aligned stochastic continuity mitigates the action mismatch induced by increasingly stale predictions in highly dynamic control.

\subsection{Ablation of EAPN Components}
We ablate EAPN on Kinetix \texttt{mjc\_swimmer}, a continuous-control task that requires
sustained motion over a relatively long horizon and is sensitive to perturbations introduced at chunk transitions.
Starting from ttRTC, we successively add Inter-Chunk Noise and Intra-Chunk Noise and compare these variants with
the complete model. We report Success Rate together with two continuity metrics.
Tail Gap is the mean $\ell_2$ deviation between the newly generated suffix after the committed prefix and
the time-aligned plan from the previous inference, characterizing consistency in the freely generated portion of
consecutive chunks.

\begin{table}[t]
\centering
\caption{Component ablation of EAPN on Kinetix \texttt{mjc\_swimmer}.}
\label{tab:eapn_ablation}
\begingroup
\setlength{\tabcolsep}{3.0pt}
\renewcommand{\arraystretch}{0.90}
\begin{tabular}{@{}lccc@{}}
\toprule
Variant & Success $\uparrow$ & Switch Gap $\downarrow$ & Tail Gap $\downarrow$ \\
\midrule
ttRTC                & 46.60 & 1.58 & 1.58 \\
+ Inter-Chunk Noise  & 76.89 & 1.55 & 1.51 \\
+ Intra-Chunk Noise  & 78.85 & \textbf{1.54} & \textbf{1.50} \\
\textbf{EAPN}        & \textbf{79.26} & 1.55 & \textbf{1.50} \\
\bottomrule
\end{tabular}
\endgroup
\end{table}

As shown in Table~\ref{tab:eapn_ablation}, Inter-Chunk Noise provides the dominant gain, raising success from 46.60 to 76.89 while reducing both continuity gaps. Adding Intra-Chunk Noise further improves success to 78.85 and yields the lowest gaps, while full EAPN reaches the best success rate of 79.26 with comparable continuity. Thus, cross-chunk correlation provides the main gain and intra-chunk correlation offers a complementary improvement.

\begin{figure}[!t]
    \centering
    \begin{minipage}[t]{0.49\linewidth}
        \centering
        \includegraphics[width=\linewidth,trim=0 45bp 0 45bp,clip]{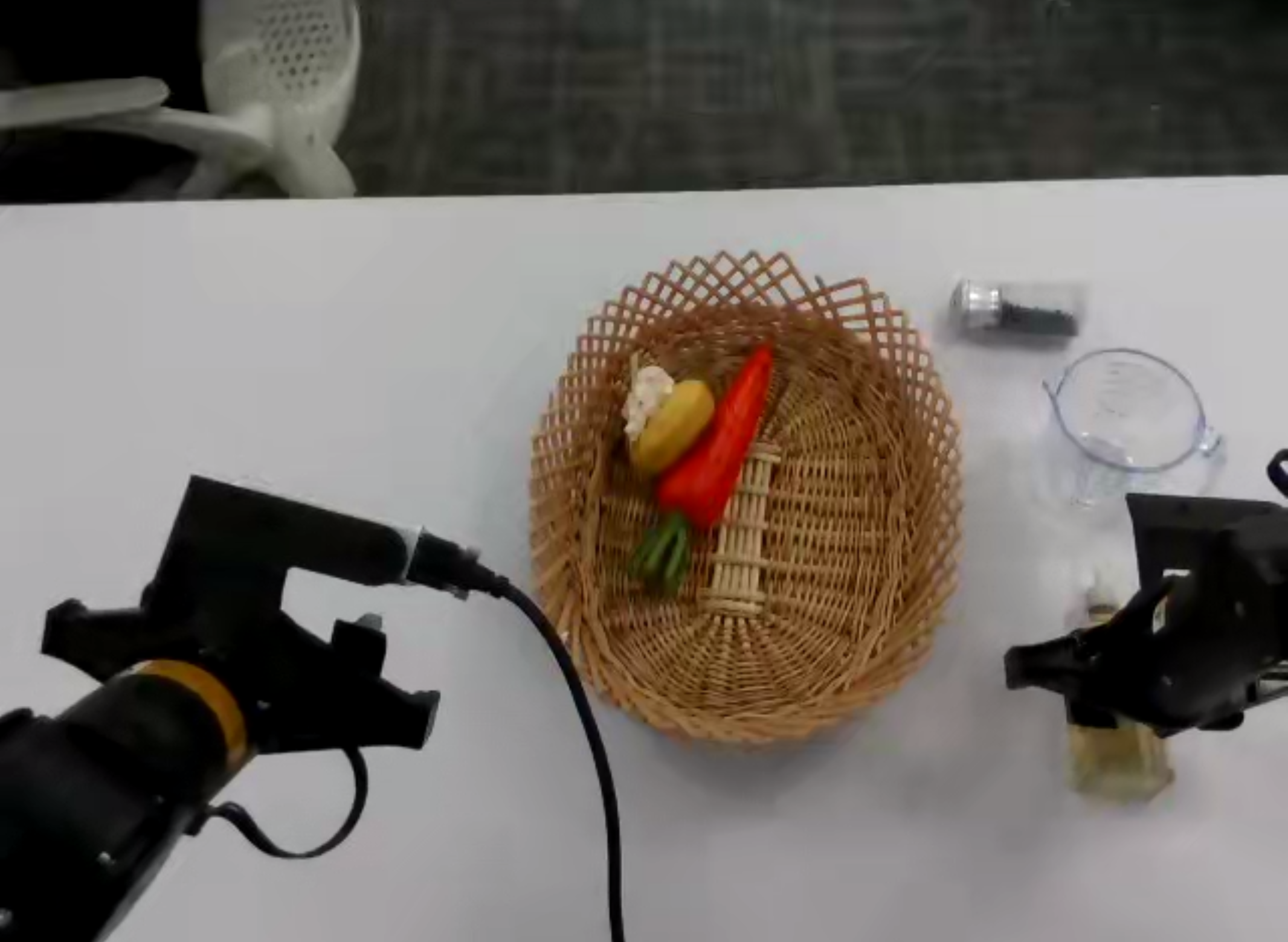}\\[-1pt]
        \footnotesize (a) Object Storage
    \end{minipage}\hfill
    \begin{minipage}[t]{0.49\linewidth}
        \centering
        \includegraphics[width=\linewidth]{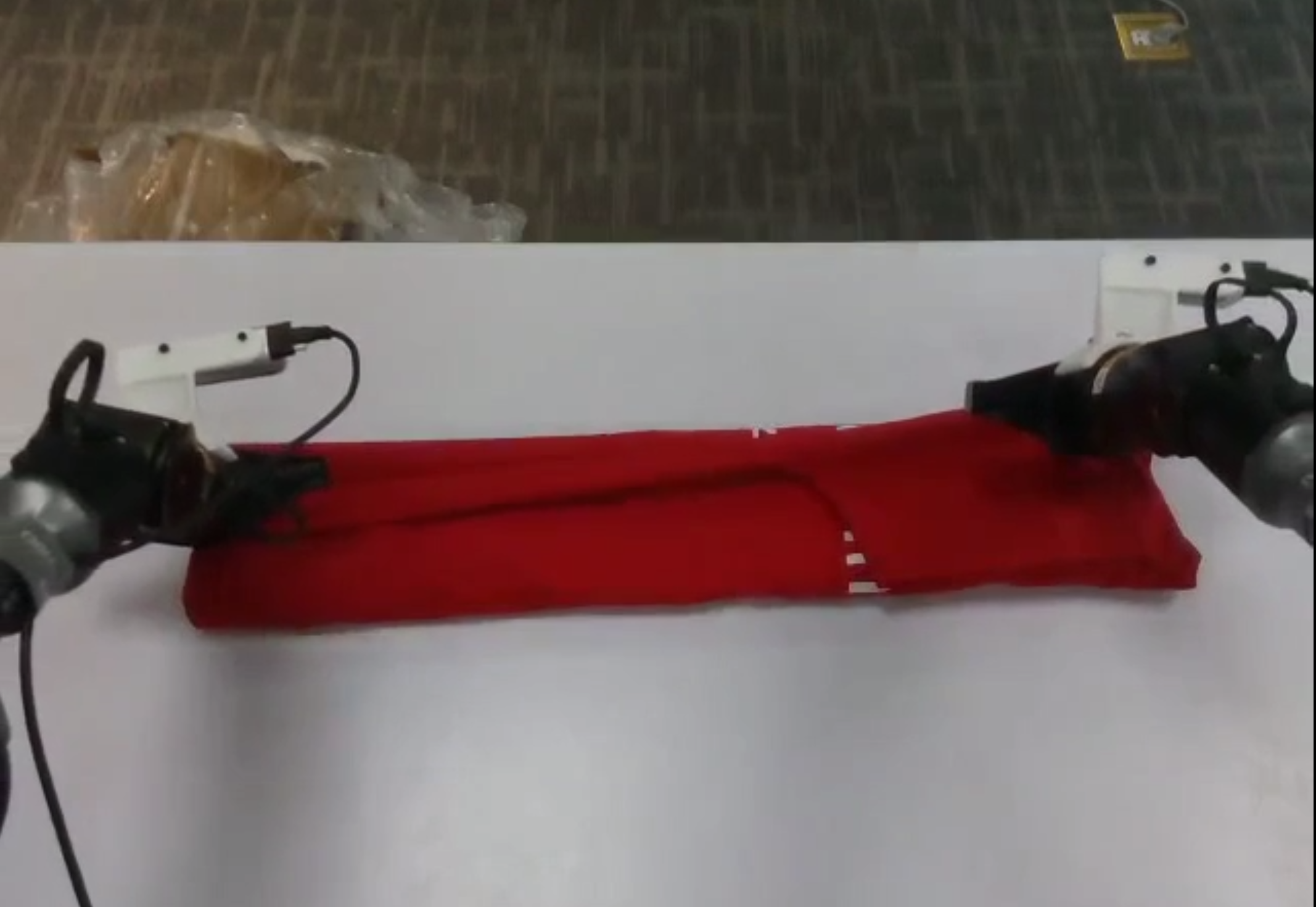}\\[-1pt]
        \footnotesize (b) Cloth Folding
    \end{minipage}
    \caption{
    Real-world experimental setups.
    (a) Object Storage;
    (b) Cloth Folding
    }
    \label{fig:real_robot_tasks}
\end{figure}

\begin{figure*}[!t]
    \centering
    \includegraphics[width=0.99\textwidth]{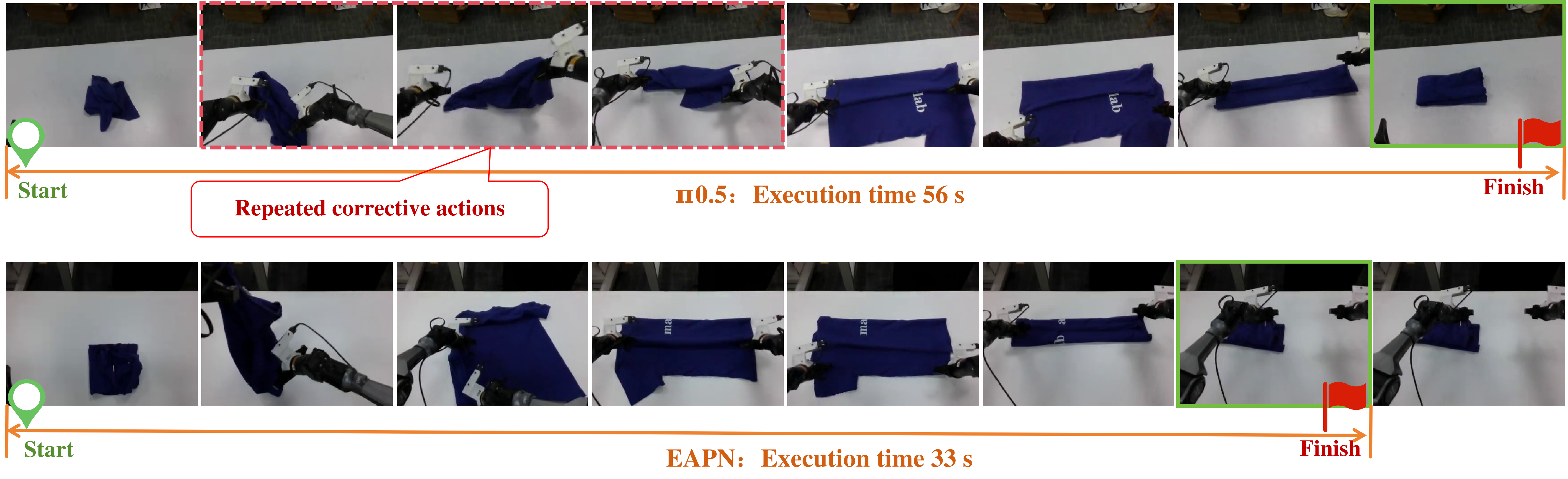}
    \caption{Representative asynchronous execution sequences for Cloth Folding.
    The baseline exhibits repeated corrective motions during execution, whereas EAPN maintains more continuous task progression in this example.}
    \label{fig:real_cloth_execution}
\end{figure*}

\subsection{LIBERO}
LIBERO comprises Spatial, Object, Goal, and Long manipulation suites~\cite{liu2023libero}. We use \piHalf~\cite{physicalintelligence2025pi05} as the base policy and evaluate two independently trained delay regimes:
\begin{equation}
    \mathcal{D}_{\mathrm{low}}=\{1,2,3,4\},\qquad
    \mathcal{D}_{\mathrm{high}}=\{5,10,15\}.
\end{equation}
Low-delay experiments use $K=5$ with training or fine-tuning matched to the small-delay range; high-delay experiments use $K=25$ and long-delay training with $D\in\{0,5,10,15,20\}$. The two groups are therefore compared only within their respective protocols and are not interpreted as a continuous delay sweep of a single model. The low-delay protocol measures performance retention under small window displacements, whereas the high-delay protocol tests robustness when the old chunk must continue executing for much longer. In both cases, the asynchronous runtime keeps consuming the current action buffer during policy inference. EAPN obtains the actual window displacement $\Delta_k$ from this buffer consumption and uses it to align the previous noise; if $E_k$ actions are executed before the next inference starts and $d_k$ more are consumed during inference, then $\Delta_k=E_k+d_k$.

\begin{table}[t]
\centering
\caption{Success rate (\%) on LIBERO under low inference delay.}
\label{tab:libero_low_delay}
\begingroup
\setlength{\tabcolsep}{3.0pt}
\renewcommand{\arraystretch}{0.90}
\begin{tabular}{@{}lcrrrrr@{}}
\toprule
Method & $D$ & Spatial & Object & Goal & Long & Avg. SR \\
\midrule
\multirow{4}{*}{VLASH} & 1 & 94.2 & 98.6 & \textbf{96.4} & 94.8 & 96.0 \\
 & 2 & 93.0 & 99.2 & 94.6 & \textbf{94.0} & 95.2 \\
 & 3 & 92.2 & \textbf{99.4} & 94.8 & 92.6 & 94.8 \\
 & 4 & 86.0 & 97.4 & 92.6 & 90.8 & 91.7 \\

\midrule
\multirow{4}{*}{ttRTC} & 1 & 94.6 & \textbf{99.0} & 94.8 & \textbf{95.4} & 96.0 \\
 & 2 & 92.4 & 98.4 & 95.0 & 93.2 & 94.8 \\
 & 3 & 92.2 & 98.0 & 96.0 & \textbf{95.0} & 95.3 \\
 & 4 & 88.8 & 95.6 & 94.2 & 90.2 & 92.2 \\
\midrule
\multirow{4}{*}{EAPN} & 1 & \textbf{97.6} & \textbf{99.0} & \textbf{96.4} & 92.6 & \textbf{96.4} \\
 & 2 & \textbf{97.8} & \textbf{99.4} & \textbf{98.0} & 92.8 & \textbf{97.0} \\
 & 3 & \textbf{95.6} & 99.2 & \textbf{96.4} & 94.4 & \textbf{96.4} \\
 & 4 & \textbf{96.4} & \textbf{99.0} & \textbf{96.6} & \textbf{94.2} & \textbf{96.6} \\
\bottomrule
\end{tabular}
\endgroup
\end{table}

\begin{table}[t]
\centering
\caption{Success rate (\%) on LIBERO under high inference delay.}
\label{tab:libero_high_delay}
\begingroup
\setlength{\tabcolsep}{3.0pt}
\renewcommand{\arraystretch}{0.90}
\begin{tabular}{@{}lcrrrrr@{}}
\toprule
Method & $D$ & Spatial & Object & Goal & Long & Avg. SR \\
\midrule
\multirow{3}{*}{VLASH} & 5 & 69.0 & 71.0 & 79.0 & 67.0 & 71.5 \\
 & 10 & 68.0 & 69.0 & 80.0 & 64.0 & 70.3 \\
 & 15 & 70.0 & 72.0 & 78.0 & 53.0 & 68.3 \\
\midrule
\multirow{3}{*}{ttRTC} & 5 & \textbf{87.0} & \textbf{96.0} & \textbf{90.0} & \textbf{86.0} & \textbf{89.8} \\
 & 10 & 76.0 & 84.0 & \textbf{91.0} & 70.0 & 80.3 \\
 & 15 & 70.0 & 80.0 & 86.0 & 56.0 & 73.0 \\
\midrule
\multirow{3}{*}{EAPN } & 5 & \textbf{87.0} & 92.0 & 89.0 & 85.0 & 88.3 \\
 & 10 & \textbf{80.0} & \textbf{88.0} & \textbf{91.0} & \textbf{73.0} & \textbf{83.0} \\
 & 15 & \textbf{72.0} & \textbf{82.0} & \textbf{88.0} & \textbf{58.0} & \textbf{75.0} \\
\bottomrule
\end{tabular}
\endgroup
\end{table}

Under low delay, EAPN maintains 96.4\%--97.0\% average success across $D=1$--$4$ and exceeds VLASH by 4.9 points at $D=4$. In the high-delay setting, ttRTC already improves substantially over VLASH, especially at $D=5$; EAPN remains competitive there and becomes stronger as the delay grows. At $D=10$ and $D=15$, EAPN reaches 83.0\% and 75.0\% average success, improving over ttRTC by 2.7 and 2.0 points, respectively. These results indicate that execution-aligned stochastic history becomes particularly useful as the prediction window grows stale and the regenerated suffix must remain consistent with a longer period of ongoing execution.

\subsection{Real-World Experiments}
We further evaluate Object Storage and bimanual Cloth Folding, which respectively test repeated rigid-object manipulation and long-horizon deformable-object coordination. Object Storage emphasizes reliable repeated grasp-and-place behavior under continuous asynchronous replanning. Cloth Folding additionally introduces nonrigid deformation, stronger historical dependence, and sustained bimanual coordination, making it more sensitive to cross-chunk inconsistency. All methods use the same \piHalf backbone, task-specific training data, and evaluation settings: 216 training episodes for Object Storage and 3,056 for Cloth Folding.

\subsubsection{Object Storage}
The robot repeatedly grasps kitchen objects of varied appearance, size, and geometry and places them into a basket; initial object positions are randomized for every rollout to reduce dependence on a fixed layout. Each method is evaluated in 20 independent 60\,s trials, and success requires storing all objects within the time limit.

\subsubsection{Cloth Folding}
Cloth Folding introduces nonrigid deformation, longer temporal dependence, and bimanual coordination. We evaluate three garments of different colors and sizes with 10 independent rollouts each, yielding 30 trials per method. The garment configuration is reset before every rollout, the execution limit is 3\,min, and a trial is counted as successful only when the target fold is completed within this limit.

\begin{table}[t]
\centering
\caption{
Comparison on real-world experiments.
}
\label{tab:real_main}
\setlength{\tabcolsep}{4.0pt}
\begin{tabular}{@{}lcc@{}}
\toprule
Method & \multicolumn{2}{c}{Success Rate $\uparrow$} \\
\cmidrule(lr){2-3}
& Object Storage & Cloth Folding \\
\midrule
Naive Async   & 65.0\% (13/20) & 63.3\% (19/30) \\
RTC           & 80.0\% (16/20) & 60.0\% (18/30) \\
ttRTC         & 80.0\% (16/20) & 63.3\% (19/30) \\
VLASH         & 85.0\% (17/20) & 93.3\% (28/30) \\
\textbf{EAPN} & \textbf{90.0\% (18/20)} & \textbf{96.7\% (29/30)} \\
\bottomrule
\end{tabular}
\end{table}

\begin{figure}[!t]
    \centering
    \includegraphics[width=\linewidth]{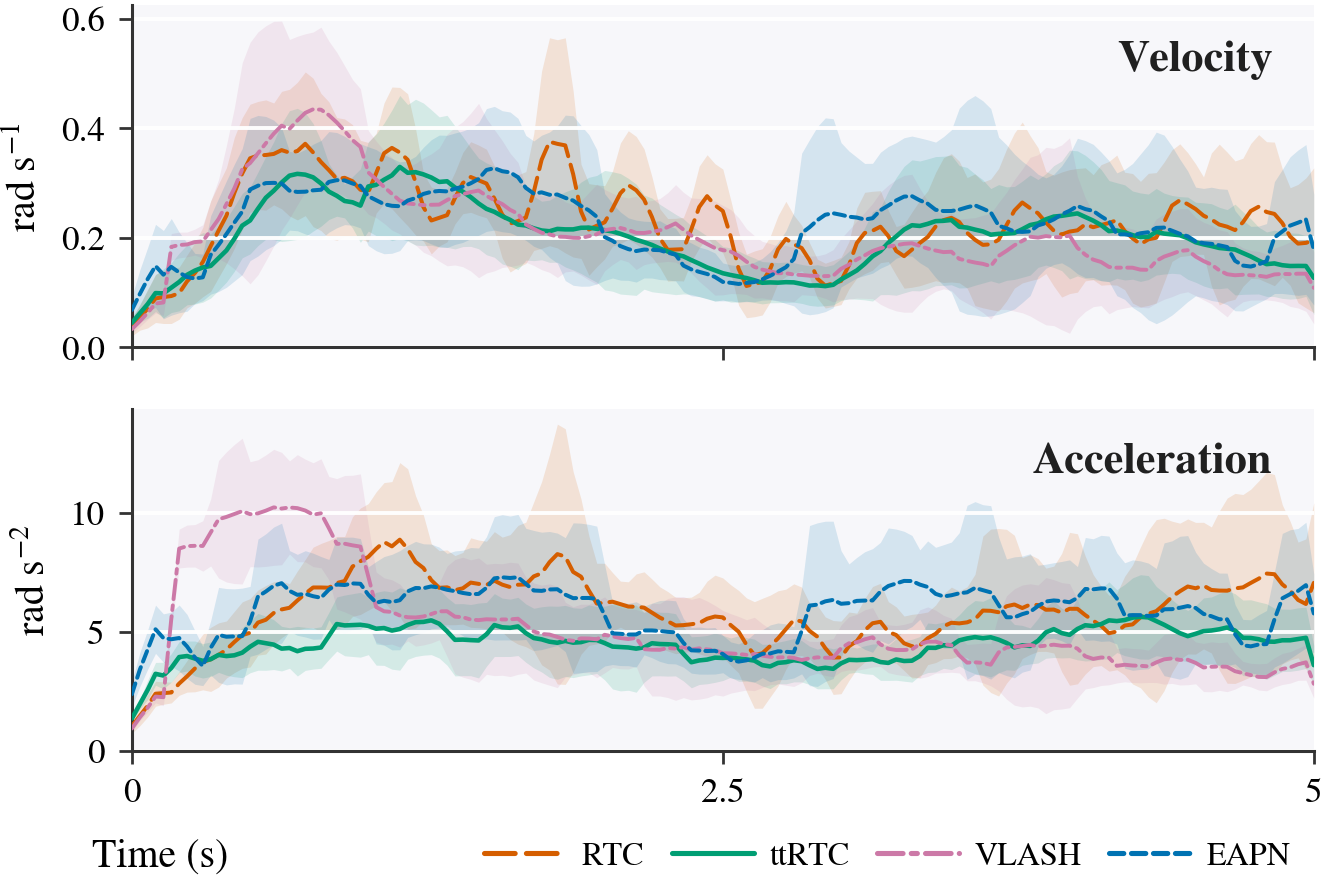}
    \caption{Motion-profile comparison on real-robot execution segments. Top: velocity. Bottom: acceleration. Curves show RTC, ttRTC, VLASH, and EAPN over 0--5\,s.}
    \label{fig:real_robot_motion}
\end{figure}

Table~\ref{tab:real_main} shows that EAPN achieves the highest success on both tasks: 90.0\% on Object Storage,
improving over Naive Async and VLASH by 25.0 and 5.0 percentage points,
and 96.7\% on Cloth Folding. In Fig.~\ref{fig:real_cloth_execution},
the baseline exhibits repeated corrections and requires 56\,s, whereas EAPN completes the task in 33\,s.
Despite using correlated stochastic states, EAPN still responds promptly to garment deformation and the latest
observation, indicating no obvious response lag.
As shown in Fig. 7, EAPN maintains velocity and acceleration profiles comparable to ttRTC,
while avoiding some of the larger fluctuations observed in RTC and VLASH.
Together with the higher task success in Table V, these results indicate that
EAPN enables more reliable continuous execution without sacrificing motion stability.

\FloatBarrier

\section{CONCLUSION}
\label{sec:conclusion}

We presented Execution-Aligned Progressive Noise (EAPN) for consistent asynchronous replanning with generative robot policies. Motivated by execution-aligned correlation in inverted source latents, EAPN models stochastic dependency across consecutive Action Chunks and temporal correlation within each chunk, using aligned historical noise as generation context. This design reduces mode switching caused by independent stochastic reinitialization without directly constraining the generated action suffix. Across D3IL, Kinetix, LIBERO, and real-robot manipulation, EAPN improves task success, cross-chunk consistency, and robustness to inference delay; Object Storage and bimanual Cloth Folding further demonstrate reliable continuous execution. These results show that execution-aligned stochastic history complements action-space temporal alignment for stable continuous generative control.

\end{document}